\documentclass[letterpaper, 10 pt, conference]{ieeeconf}  % Comment this line out if you need a4paper

\IEEEoverridecommandlockouts                              % This command is only needed if 
\usepackage{graphics} % for pdf, bitmapped graphics files
\usepackage{epsfig} % for postscript graphics files

\usepackage{color}
\usepackage{url}
\usepackage{hyperref}
\usepackage{amsmath}
\usepackage{relsize}

\definecolor{patrick_color}{rgb}{.6,.4,.05}
\definecolor{chengcheng_color}{rgb}{.5,.7,.1}
\definecolor{chris_color}{rgb}{0,0.35,0}
\definecolor{chengde_color}{rgb}{0,0,1}
\definecolor{charlie_color}{rgb}{0,0,0.8}
\definecolor{samarth_color}{rgb}{0.75,0.25,0.0}
\definecolor{jeremy_color}{rgb}{0,.7,.7}

\newcommand{\method}[0]{VPEC}
\newcommand{\network}[0]{VPEC-Net}

\title{\LARGE \bf Visual Pressure Estimation and Control for Soft Robotic Grippers}

\author{Patrick Grady$^{1}$, Jeremy A. Collins$^{1}$, Samarth Brahmbhatt$^{2}$, Christopher D. Twigg$^{3}$,\\Chengcheng Tang$^{3}$, James Hays$^{1}$, Charles C. Kemp$^{1}$% <-this % stops a space
\footnotemark{}
}

\newenvironment{first_caption}
  {\par\footnotesize}
  {\par\addvspace{\bigskipamount}}

\begin{document}

\twocolumn[{%
\renewcommand\twocolumn[1][]{#1}%
\maketitle
\thispagestyle{empty}
\pagestyle{empty}

\begin{center}
\centering
\vspace{-2mm}
\includegraphics[width=1.0\linewidth]{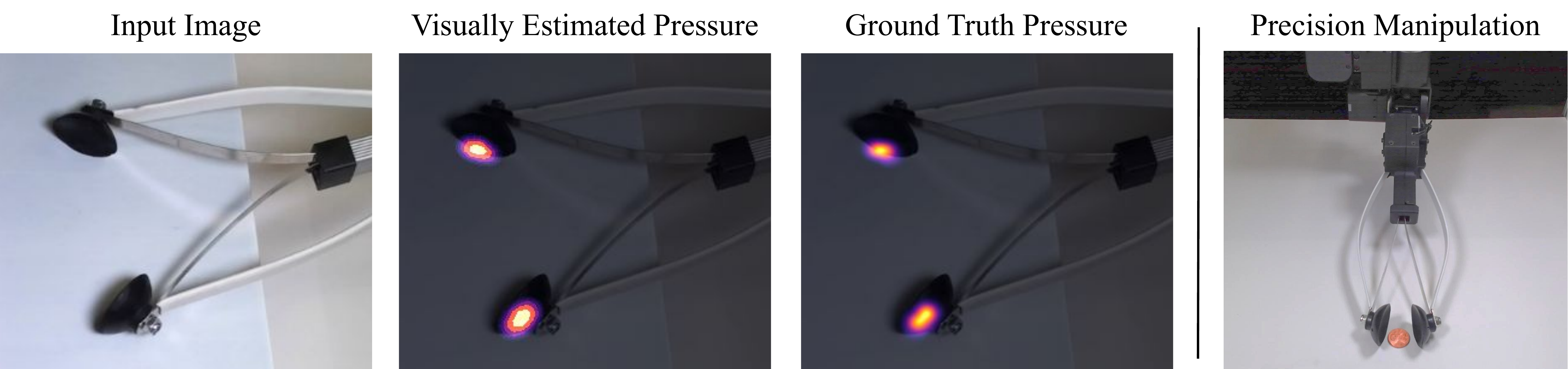}
\end{center}
\vspace{0mm}
\begin{first_caption}
 Fig. 1. \textbf{Left}: Given a single RGB input image, \network{} estimates the pressure applied by a soft gripper to a flat surface. \textbf{Middle Left and Middle Right}: Pressure images are shown overlaid on the input image. \network{} outputs a pressure image with a pressure estimate for each pixel of the input image. The estimated pressure image closely matches the ground truth pressure image obtained with a planar pressure sensing array. \textbf{Right}: Using visual servoing with estimated pressure images, a robot grasps small objects, including a penny.
\end{first_caption}
%\vspace{3mm}
}]
\setcounter{figure}{1}  % start figure numbers at 2
\setcounter{footnote}{1}

% \thanks{test2}

% \maketitle
% \thispagestyle{empty}
% \pagestyle{empty}

% \begin{abstract}
% \textbf{\textit{Abstract---} Soft robotic grippers facilitate contact-rich manipulation, including robust grasping of varied objects. Yet the beneficial compliance of a soft gripper also results in significant deformation that can make precision manipulation challenging. We present visual pressure estimation \& control (\method{}), a method that uses a single RGB image of an unmodified soft gripper from an external camera to directly infer pressure applied to the world by the gripper. We present inference results for a pneumatic gripper and a tendon-actuated gripper making contact with a flat surface. We also show that \method{} enables precision manipulation via closed-loop control of inferred pressure. We present results for a mobile manipulator (Stretch RE1 from Hello Robot) using visual servoing to do the following: achieve target pressures when making contact; follow a spatial pressure trajectory; and grasp small objects, including a microSD card, a washer, a penny, and a pill. Overall, our results show that \method{} enables grippers with high compliance to perform precision manipulation.}
% \end{abstract}

\footnotetext{Patrick Grady, Jeremy A. Collins, James Hays, and Charles C. Kemp are with the Institute for Robotics and Intelligent Machines at the Georgia Institute of Technology (GT). $^2$Samarth Brahmbhatt is with Intel Labs. $^3$Christopher D. Twigg and Chengcheng Tang are with Meta Reality Labs. This work was supported in part by NSF Award \# 2024444. Code, data, and models are available at \url{https://github.com/Healthcare-Robotics/VPEC}. Charles C. Kemp is an associate professor at GT. He also owns equity in and works part-time for Hello Robot Inc., which sells the Stretch RE1. He receives royalties from GT for sales of the Stretch RE1.}

\begin{abstract}
Soft robotic grippers facilitate contact-rich manipulation, including robust grasping of varied objects. Yet the beneficial compliance of a soft gripper also results in significant deformation that can make precision manipulation challenging. We present visual pressure estimation \& control (\method{}), a method that infers pressure applied by a soft gripper using an RGB image from an external camera. We provide results for visual pressure inference when a pneumatic gripper and a tendon-actuated gripper make contact with a flat surface. We also show that \method{} enables precision manipulation via closed-loop control of inferred pressure images. In our evaluation, a mobile manipulator (Stretch RE1 from Hello Robot) uses visual servoing to make contact at a desired pressure; follow a spatial pressure trajectory; and grasp small low-profile objects, including a microSD card, a penny, and a pill. Overall, our results show that visual estimates of applied pressure can enable a soft gripper to perform precision manipulation.

\end{abstract}

\section{Introduction}

High compliance helps soft robotic grippers conform to the environment and apply low forces during contact, but it also results in significant deformation that makes precise motions more difficult to achieve. For precision manipulation tasks, small position errors due to deformation can lead to failure. For example, tasks using fingertips to press a small button, flip a small switch, or pick up a small object depend on pressure being applied with precision. 

One approach to precisely apply pressure is to explicitly model the mechanics of the gripper. Rigid-body models enable precise control of rigid grippers, but result in errors when applied to soft grippers. For example, sliding a soft gripper's fingertips across a surface can result in large deviations from the gripper's undeformed geometry. Soft-body models can represent the compliant geometry of soft grippers, but are more complex than rigid-body models and depend on quantities that can be difficult to measure, such as external forces and internal strain. Embedded sensors in soft grippers can make relevant measurements, but increase hardware complexity and often alter gripper mechanics. For both rigid and soft grippers, an explicit model of the gripper needs to be related to the environment to model applied pressure, which often involves additional sensors and calibration.  
% Soft-body models can represent the compliant nature of soft grippers. However, these models are often complex and depend on additional information that is difficult to measure, such as contact forces. Higher accuracy kinematic estimates can be achieved by embedding sensors that measure quantities such as external loads, internal strain, and contact locations. Yet, additional sensors increase system complexity and alter the mechanics of the soft gripper. Moreover, even a precise model of gripper geometry and applied loads typically needs to be related to manipulation objectives defined with respect to the world. Additional sensors and calibration can be used to explicitly relate the gripper's geometric model to the world, but adds further complexity. 

We present a novel approach that circumvents these modeling and instrumentation complexities by using an external camera to directly estimate the pressure applied by a soft gripper to the world. Our approach relies on two key insights. First, many manipulation tasks only depend on the pressure applied by the gripper to the world, rather than the gripper's detailed state. For these tasks, directly estimating and controlling pressure applied to the world is sufficient for task success. Second, the pressure applied to the world by a soft gripper can be directly estimated by the gripper's visible deformation. This takes advantage of high compliance, since larger deformations are more easily observed by an external camera. 

Our method, visual pressure estimation \& control (\method{}), uses a convolutional neural network, \network{}, to infer a 2D \textit{pressure map} overlaid on the input RGB image from an external camera (see Figure 1). In other words, contact locations and pressure are estimated \emph{in the image space} with an estimated pressure for each pixel in the image. A control loop achieves pressure objectives in the image space, enabling a robot to precisely control pressure applied to the world and thereby grasp a small object observed by the camera. In addition to gripper deformation, \network{} has the potential to use other information, such as cast shadows and motion blur. 

For this paper, we consider contact with a horizontal plane, which is a common surface relevant to manipulation. To construct a training dataset, we hand-operated a tendon-driven soft gripper and a pneumatic soft gripper to make contact with a high-resolution planar pressure sensor. We capture these interactions with four RGB cameras and use camera extrinsics to project the pressure sensor data onto the RGB images, creating a labeled dataset for training \network{}. We collected approximately one hour of data for each gripper, yielding a dataset of 650K frames. Our contributions include the following:

\begin{itemize}
    \item \textbf{\method{}}: An algorithm that infers pressure applied by a soft gripper to a planar surface using a single RGB image.
    % \item We evaluate \method{} on an unseen split of the dataset and ``in the wild''.
    \item \textbf{Precision Manipulation with VPEC:} Evaluations in which a mobile manipulator with a soft gripper achieves pressure objectives via closed-loop control and grasps small objects, including a washer and a coin. 
    \item \textbf{Release of dataset, trained models, and code:} We will release our core methods online to support replication of our work. 
\end{itemize}

%The output of \network{} is used for closed-loop, image-based visual servoing to grasp small object while regulating the pressure applied to the surface.
% \input{1_introduction}
\section{Related Work}

Our work builds on prior efforts to visually infer pressure applied by human hands \cite{grady2022pressurevision}. We use the same neural network architecture, but apply it to inference and control of soft robotic grippers. 

We evaluate our approach with the task of precision manipulation on a flat surface. Our method controls the pressure applied by a soft gripper to the surface to grasp small objects. Similarly, humans often slide their fingertips across flat surfaces when picking up small objects \cite{kazemi2014human, eppner2015exploitation}, which has inspired robotic grasping methods \cite{ciocarlie2014velo, babin2018picking, yoon2021analysis}. Work on grasping with soft end effectors has focused on larger objects than we consider \cite{eppner2017visual, pozzi2020hand, gupta2016learning}. Grasping smaller objects tends to be more sensitive to gripper deformation, such as deformation due to sliding while in contact. %\jeremy{less sensitive to geometric variation from deformation due to contact with the object or its immediate environment.} 

% internal cameras for proprioception and tactile sensing
% she2020exoskeleton

% original gelsight paper, but not integrated on a gripper?
% johnson2009gelsight

%, belzile2016stiffness, della2017estimating, abdeetedal2018grasp
% lin2009signal,  

% , liang2021contact
% she2020exoskeleton,

Prior work has used internal sensors to infer contact and pressure based on deformation of the gripper's surface \cite{begej1988planar, li2014localization, yamaguchi2016combining, kuppuswamy2020soft, lepora2021soft}, changes to gripper vibration \cite{kuang2020vibration}, deflection of the gripper's compliant joints \cite{koonjul2011measuring}, and changes to the gripper's motion \cite{wang2020contact}. We expect that our method can use similar information by observing a soft gripper with an external camera. For this paper, estimation only uses a single image, which can have motion blur but lacks information available in sequences of images. 

% kennedy2005vision, noohi2014using, shin2019sequential, marban2019recurrent,
% greminger2003modeling
% greminger2004vision

Research on image-based force estimation has focused on inferring force and torque applied by a rigid tool to a deformable object \cite{nazari2021image, kennedy2005vision, noohi2014using, kim2019efficient, marban2019recurrent, chua2021toward}. Early work inferred grip force for a microgripper \cite{greminger2003modeling}. Cross-modal research has used vision to predict the output of robot-mounted tactile sensors \cite{li2019connecting, zapata2020generation, patel2020deep}. Our approach relies on soft grippers that visibly deform to infer the output of tactile sensors mounted to the world. 

% zapata2020generation

% \ck{add citations here?} \cite{}

%Kinematic models are often used to transform a gripper's internal contact and pressure sensor readings to the surrounding environment, but deformations due to high compliance can be difficult to model. Our method operates directly in the image space of a camera viewing the gripper and the environment. 

Our method uses visual servoing \cite{hutchinson1996tutorial} to achieve pressure objectives in the camera's image space. Prior work has demonstrated visual control of a robot's arm relative to flat surfaces based on shadows \cite{fitzpatrick2004power}. Marker-based visual servoing has achieved precise in-hand manipulation with a soft gripper \cite{calli2018robust}. Visual object tracking has enabled precision insertions with a soft gripper \cite{morgan2021vision}. Our system enables a soft gripper to grasp small, low-profile objects from a flat surface.

%Even with rigid robots, uncalibrated, marker-based visual servoing can outperform joint control for precision manipulation \cite{jagersand1997experimental}, and markerless visual servoing has achieved micro-scale precision \cite{onal2007visual}.

Our system can be thought of as using a \textit{virtual} tactile sensor array mounted to the world with which it attempts to achieve pressure objectives. As such, our approach is similar to tactile servoing with real tactile sensor arrays \cite{sikka1994tactile, chen1995edge, li2013control, wen2021tactile, lepora2021pose}. Notably, our system reports inferred pressure for each pixel of the input RGB image. This enables our system to directly relate pressure and vision. For example, in our grasping evaluation, the robot uses the RGB image to find the centroid of the target object, which determines key pressure objectives.

\section{Visual Pressure Estimation}

In this section, we describe the grippers used and the capture process to create our dataset. We also describe the network architecture and training procedure of  \network{}.

\begin{figure*}
  \centering
  \vspace{1mm}
  \includegraphics[width=0.95\linewidth]{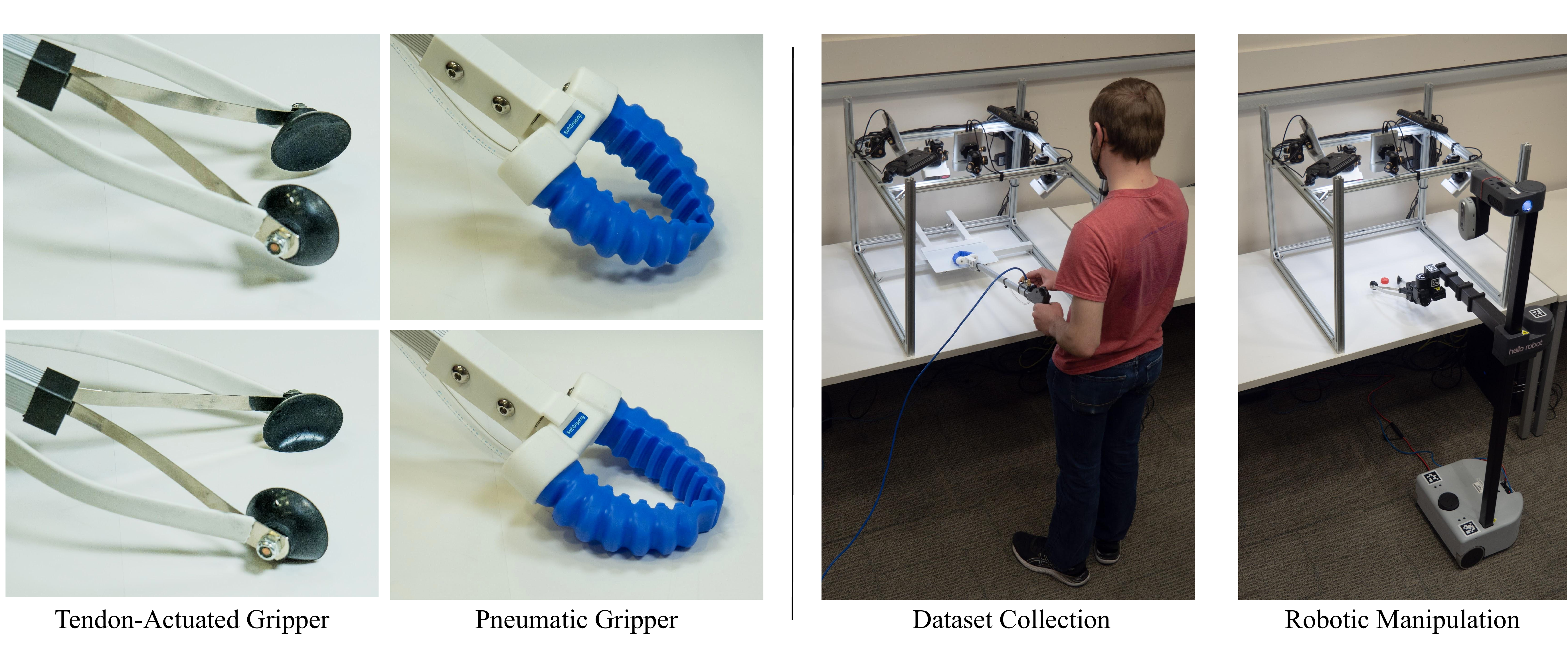}
  \vspace{-1mm}
  \caption{\textbf{Left}: The two grippers used to train and test \network{}.  In the top images, the grippers are hovered in mid-air, and in the bottom images, the grippers are pressed against the surface. Notice the deflection in the pneumatic gripper and the deformation in the tips of the tendon-actuated gripper. \textbf{Right}: We use hand-operated grippers to make contact with a high-resolution pressure sensing array to collect training data. During robotic manipulation experiments, we command a Hello Robot Stretch RE1 robot to pick up a variety of small objects from the table.}
  \label{fig:camera_cage}
\end{figure*}

\subsection{Selected Grippers}

For a vision-based approach to successfully estimate contact and pressure, there must be visual cues to indicate the presence of these quantities. As a result, we train and test \network{} with soft robotic grippers. These grippers are compliant and deform when in contact with an object. We select two different models of grippers that are examples from common classes of grippers used by researchers. %Two types of grippers were selected: a tendon-actuated soft gripper and a pneumatic soft gripper (Figure X).

\textbf{Tendon-Actuated Gripper}: The first gripper we consider is the \textit{Stretch Compliant Gripper} that comes with the \textit{Stretch RE1} mobile manipulator by Hello Robot Inc. This gripper has suction-cup-shaped soft rubber fingertips supported by spring steel flexures that bend when the gripper closes. To close the gripper, an actuator uses a tendon to pull the inner flexures.  In a user study, a similar commercially available grabber tool was found to be adept at manipulating various household objects~\cite{kemp2021design}. 
During the collection of the dataset (Sec \ref{sec:data_capture}), we used a hand-operated version of this gripper.

% During data collection, we use a hand-operated version.

%We evaluate the gripper's force profile in Figure X. The clamping force is dependent both on the gripper actuation percentage and the width of the object being grasped. We find that when grasping an object 20mm thick, the gripper can exert a maximum of 30 N of gripping force.

The gripper displays several visual cues that may indicate pressure when grasping. The rubber fingertips visibly deform to match the contours of the grasped object or when in contact with a surface, and the steel flexures bend when in contact with a surface (Figure \ref{fig:camera_cage}).

% \subsubsection{Pneumatic Gripper}
\textbf{Pneumatic Gripper}: The second gripper we consider is a pneumatic gripper sold by \textit{SoftGripping GmbH} \cite{softgripping}. The gripper is made of a flexible silicone, and contains hollow cavities for pressurized air. When inflated, the sides of the finger expand asymmetrically, resulting in the fingers closing.

%This type of gripper is currently under active research in the robotics community.

% The pneumatic gripper features a hollow cavity designed to be filled with compressed air. As the pressure inside the cavity increases, one side of the finger stretches while the other side is relatively inelastic, causing a bending of the finger. In the center of each finger is a cavity which can be filled with compressed air, in the range of -50 to +120 kPa gauge pressure. 
% We find that the gripper can exert up to 2.5 N of gripping force when grasping a 20mm thick object.

Due to its silicone construction, the gripper is soft, and the entire finger deforms when in contact (Figure \ref{fig:camera_cage}). Additionally, as the pressure in the finger cavity increases, the exterior of the gripper expands.

% \subsubsection{Rigid Gripper} The final gripper used is a relatively rigid design. It was selected due to its stiffness, and it serves to evaluate VPE's ability to estimate pressure for non-soft grippers that do not conform to their environment and display few visual cues to indicate pressure.

% The rigid gripper is a commercially available grabber design. It is made of rigid plastic and uses stiff rubber inserts in the jaws which have minimal deflection during grasping. %\patrick{Ettore Grip 'n Grab}

% As the rigid gripper grasps an object, there is minimal visible deformation in the fingers once contact is made. While a small amount of bending is visible near the jaw hinge points, and in the rubber pads of the fingertips, this is relatively small as compared to the other grippers (Figure X).

\subsection{Data Capture Setup} \label{sec:data_capture}

We built a custom data capture rig to collect RGB images with synchronized ground truth pressure.  
The rig uses aluminum framing to rigidly support a pressure sensor and cameras. The parts of the rig visible to cameras are covered with a white vinyl covering to provide a consistent visual background.

To record pressure data, we use a Sensel Morph \cite{senselmorph} sensor. The Morph is a flat pressure sensor with an active area of $23\times13$ cm and features a grid of $185\times105$ individual force-sensitive resistor (FSR) elements. The sensor produces high-resolution pressure data at approximately 100 Hz. %We find the maximum pressure value in the dataset is 150 kPa \jeremy{Wouldn't this also just be max pressure of the Sensel?}.

Four Azure Kinect cameras are mounted at different locations around the cage to observe the pressure sensor and gripper from a variety of viewpoints. The RGB feed from the cameras is captured in 1080p at 30 Hz. The capture rig additionally has two lights mounted to provide illumination which can be turned on or off in any combination. Bright lighting reduces the effect of motion blur, but during fast motions, some blurring is still visible.

The cameras and pressure sensor are calibrated before each recording session using a specialized fiducial board. The board uses ChArUco \cite{garrido2014automatic} markers on the top for localization in camera space, while pins on the bottom push into the edges of the pressure sensor,
% constraining the transform from the 2D pressure sensing grid to image space.
allowing consistent positioning.

\subsection{Data Capture Protocol}

While our work is targeted toward robotic grippers, we operate the grippers by hand for data collection (Figure \ref{fig:camera_cage}). Collecting data with a human operator allows for efficient capture of a diverse dataset including a wide range of pressure levels, orientations, grasp styles, and speeds. The grippers are mounted on a handle $60$ cm in length to allow a person to operate the gripper easily. In Section \ref{sec:robotic_system}, we show that a network trained on this data can be used to control the position of a robot-actuated gripper.

We designed a capture protocol to systematically collect data from the grippers. We studied actions where the gripper makes contact with the surface of the planar pressure sensor. Our data is divided into three classes of actions: \textit{make contact}, where both fingers of the gripper are lowered onto the surface, \textit{slide}, where the gripper is translated along the surface, and \textit{close gripper}, where the gripper is closed while in contact with the surface. We additionally collect \textit{no contact} data, where the gripper held just above the surface without making contact to provide adversarial training data.

Data collection is further divided by the amount of force applied, the lighting configuration, the speed of the human operator, and the approach angle of the gripper. At least 30 seconds of data is collected for each combination of parameters, where the operator approaches the sensor and performs multiple grasps. Between individual grasps, the operator varied the translation and angle of the gripper. We record 32 actions each with 3 lighting conditions, resulting in 96 individual sequences for each gripper. Approximately 1 hour of data is collected for each gripper.

We randomly remove $20\%$ of the sequences to create a held-out test set.

\begin{figure*}
  \centering
  \vspace{2mm}
  \includegraphics[width=1.0\linewidth]{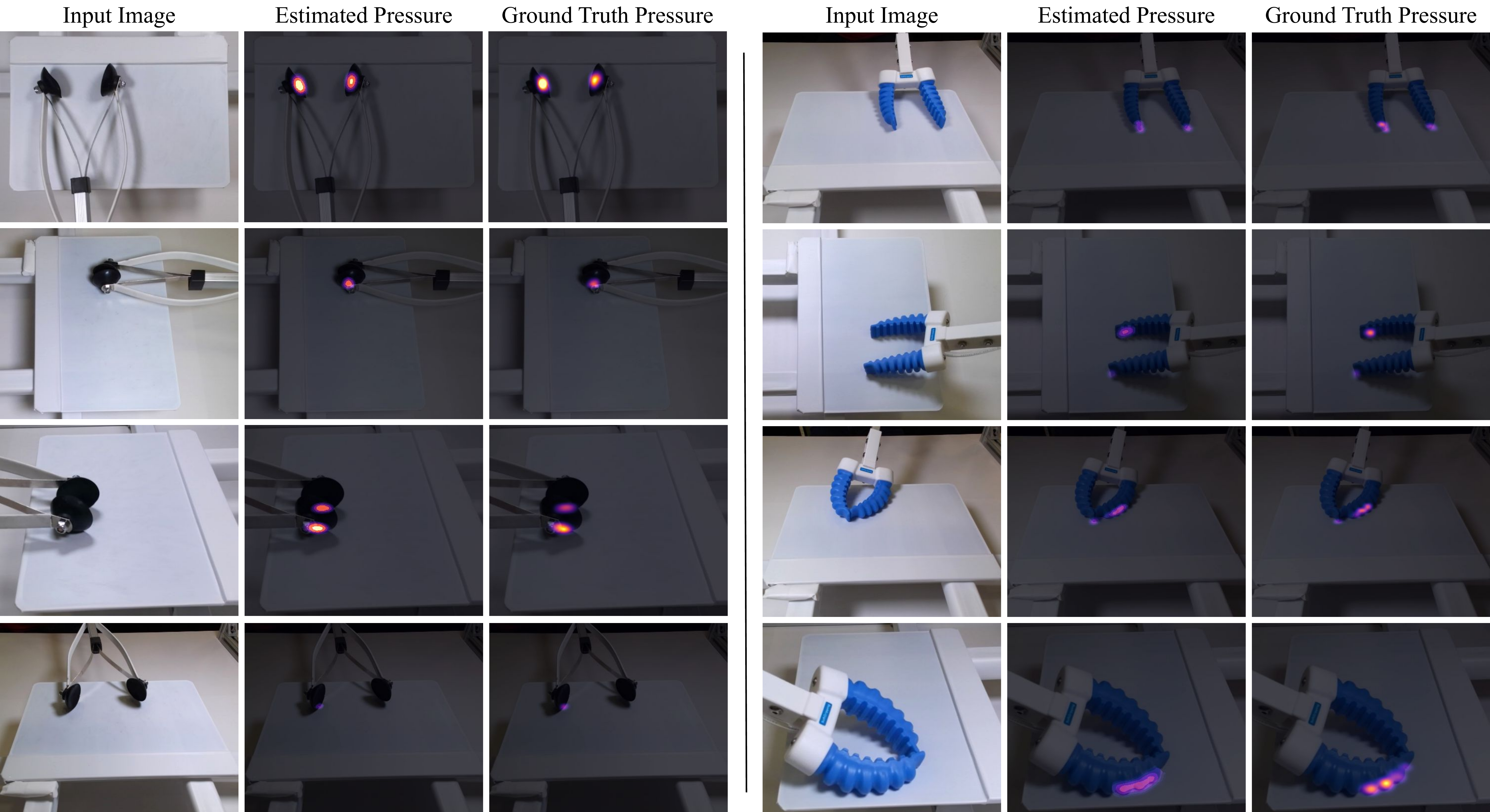}
  \vspace{-4mm}
  \caption{\textbf{Left}: Examples of pressure estimation for the tendon-actuated gripper. \textbf{Right}: Examples of pressure estimation for the pneumatic gripper. \textbf{Input Image Column}: The input RGB image used by \network{} to infer pressure. \textbf{Estimated Pressure Column}: The estimated pressure image overlaid on the input image. \textbf{Ground Truth Pressure Column}: The ground truth pressure measurements from a pressure sensing array overlaid on the input image.}
  \label{fig:network_gt_comparison}
\end{figure*}

\begin{table*}[h]
\centering
% use resizebox if table gets too big, also uncomment the } after \end{tabular}
%\resizebox{\textwidth}{!}{
\begin{tabular}{c|c|c|c|c}
    \textbf{Method} & \textbf{Temporal Acc.} & \textbf{Contact IoU} & \textbf{Volumetric IoU} & \textbf{MAE}\\\hline
    % Stretch - all actions & 95.1\% & 66.7\% & 54.0\% &  6.2 Pa \\\hline
    Tendon-Actuated Gripper & 95.9\% & 73.8\% & 58.2\% &  5.3 Pa \\\hline
    % Soft Gripper - all actions & 94.4\% & 61.0\% & 50.1\% & 9.6 Pa \\\hline
    Pneumatic Gripper & 95.1\% & 63.3\% & 52.0\% & 9.7 Pa \\\hline
    % Rigid Gripper & 80.7\% & 50.3\% & 38.0\% & 12.0 Pa \\\hline
\end{tabular}
%}
\caption{Results of Visual Pressure Estimation}
\label{tab:network_results}
\end{table*}

\subsection{Network Architecture}

We develop \network{}, a neural network to estimate pressure in \textit{image space}. The network uses a single RGB image as input and produces an estimated pressure for each input pixel. To generate ground truth pressure data for this approach, the data measured by the pressure sensor is warped into image space using a homography transform. This allows directly overlaying pressure information onto the image.

\network{}'s architecture takes inspiration from image-to-image translation neural networks used in the semantic segmentation literature. For an input RGB image $I$, a pressure image $\hat{P}=f(I)$ is estimated. The network uses an encoder-decoder architecture with skip connections. An SE-ResNeXt50 network
\cite{resnet, squeeze-excitation, resnext, segmentation_models_pytorch}
with weights from pretraining on ImageNet \cite{deng2009imagenet} is used for the encoder, and an FPN network \cite{fpn} is used for the decoder.

The task of pressure estimation is framed as a classification problem. The pressure range is divided into 8 discrete bins placed evenly in logarithmic space, including an additional \textit{zero pressure} bin. For each pixel in the output pressure image, the network classifies which pressure bin the pixel should reside in. The network is trained with a cross-entropy loss. We tested various output representations and found that this outperformed a direct regression of a pressure scalar.

Images from the cameras are cropped to extend slightly past the edges of the pressure sensor. \network{} is trained for 600k iterations using the Adam optimizer \cite{adam}. The learning rate is initially set at $1\mathrm{e}{-3}$, which drops to $1\mathrm{e}{-4}$ after 100k iterations. During training, several types of augmentations are used, including flips, random rotations, translations, scaling, brightness, and contrast changes.

\section{Evaluation of Visual Pressure Estimation}

To evaluate the performance of \network{}, we perform evaluations on the held-out test set. We use a variety of evaluation metrics similar to \cite{grady2022pressurevision} to quantify pressure estimation accuracy.

\paragraph{Temporal Accuracy} To evaluate the temporal accuracy of pressure estimates, if \textit{any} pressure pixel is above a threshold of 1.0 kPa, the frame is marked as containing contact. Temporal Accuracy measures the consistency between the presence of ground truth and estimated contact. %\jeremy{any justification for the 0.5 kPa threshold? How does the reader know this number isn't cherrypicked?}

% \begin{equation}
%     Acc_{temp}=\frac{TP+TN}{TP+FP+TN+FN}
% \end{equation}

% We use the intersection over union of binary contact images t

\paragraph{Contact IoU} To determine the spatial and temporal accuracy of pressure estimates, binary contact images are generated by thresholding pressure at the same value used for \textit{temporal accuracy}. The ground truth contact image and estimated contact image are compared to calculate intersection over union (IoU).

\paragraph{Volumetric IoU} To assess the magnitude of pressure estimates, we extend the Contact IoU to Volumetric IoU. This views pressure images as 3D volumes, with the height of the volume proportional to the quantity of pressure. The metric calculates the intersection over union of the two volumes and returns a percentage.

% \jeremy{An IoU visualization similar to the PressureVision paper might look good here}

\begin{equation}
    IoU_{vol}=\frac{\sum^{i,j}min(P_{i,j}, \hat{P}_{i,j})}{\sum^{i,j}max(P_{i,j}, \hat{P}_{i,j})}
\end{equation}

\paragraph{MAE} To quantify the error in pressure in physical units, the mean absolute error is calculated across each pixel. As most of the pixels in the dataset contain zero pressure, the MAE is low compared to the peak pressure observed in the dataset.

\subsection{Results} We train one network for each gripper and measure performance on the held-out test set. The results are reported in Table \ref{tab:network_results}. We also provide qualitative examples of the network pressure prediction in Figure \ref{fig:network_gt_comparison}.

Generally, \network{} performs well at estimating pressure from a single image. Our approach can accurately estimate if the gripper is in contact with the surface or not, achieving a temporal accuracy $>95\%$ for both grippers. 

We observe that the network trained on the tendon-actuated gripper outperforms the pneumatic gripper in all metrics. We hypothesize that this is because the shape of the pressure distribution created by the tendon-actuated gripper is often simpler (Figure \ref{fig:network_gt_comparison}). The tendon-actuated gripper also tends to visibly deform in a localized way, while deformation of the pneumatic gripper is less local and occurs across a wide area. 

\subsection{Limitations} While the network successfully reconstructs pressure on a flat surface, our dataset does not include objects with curved surfaces or unseen textures and only includes a limited set of action classes. 

%This as a limitation of our method, and we leave the capture of more diverse data to future work.

\begin{figure}[t]
  \centering
  \vspace{2mm}
  \includegraphics[width=1.0\linewidth]{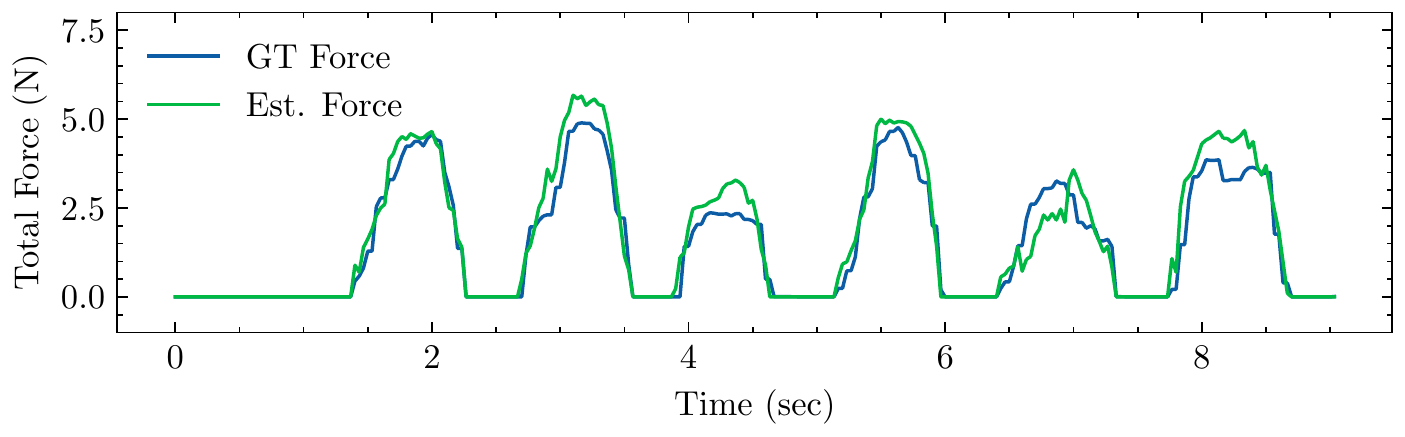}
  \includegraphics[width=1.0\linewidth]{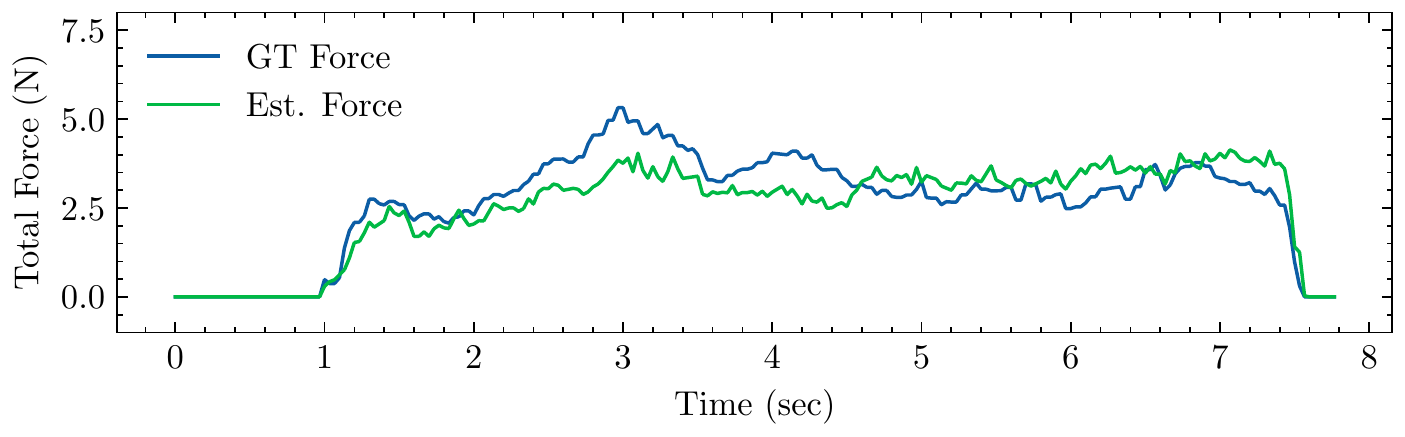}
  \vspace{-5mm}
  \caption{Force estimates over time are visualized for \textit{make contact} and \textit{slide} sequences of the tendon-actuated gripper in the test set. While \network{} displays some amount of error in the quantity of force exerted, it accurately captures the onset and termination of contact.}
  \label{fig:force_over_time}
%   \vspace{-2mm}
\end{figure}

% \patrick{Do camera generalization: train on three, test on last one? Generate saliency images similar to the pressurevision paper?}

\begin{figure}[b]
  \centering
  \includegraphics[width=0.9\linewidth]{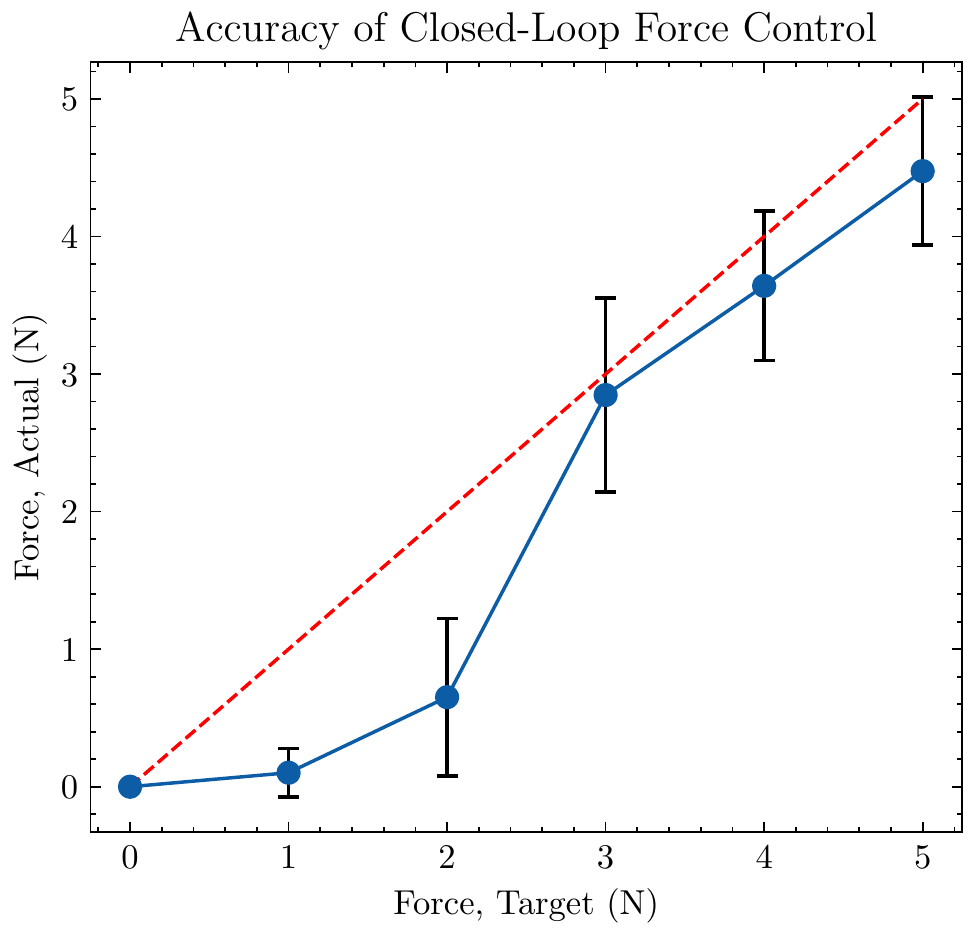}
  \vspace{-2mm}
  \caption{We use a simple controller to achieve a target force, applying feedback from \network{}'s visual pressure estimation.  The actual force is measured using the pressure sensor and matches the target value well at higher force levels.
  \label{fig:force_actual}}
\end{figure}

\begin{figure*}
\begin{minipage}[t]{0.31\textwidth}
  \centering
  % \vspace{1mm} 
  \vspace{6.5mm} 
  \includegraphics[width=1.0\linewidth]{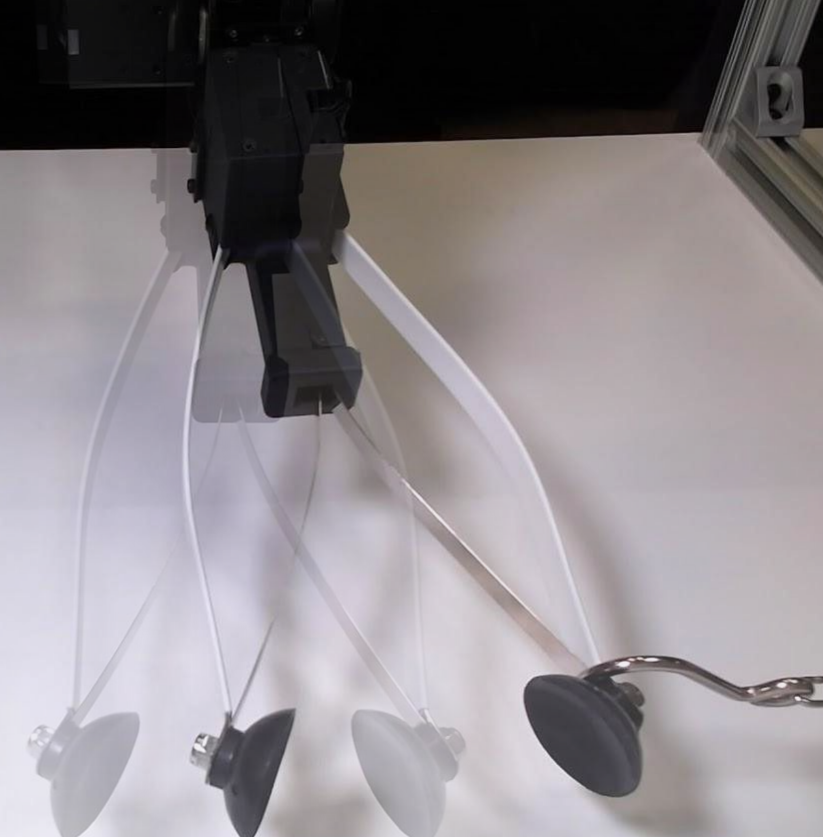}
  \vspace{0mm}
\end{minipage}
\hfill
\begin{minipage}[t]{0.38\textwidth}
  \centering
  \vspace{2mm}
  \includegraphics[width=1.0\linewidth]{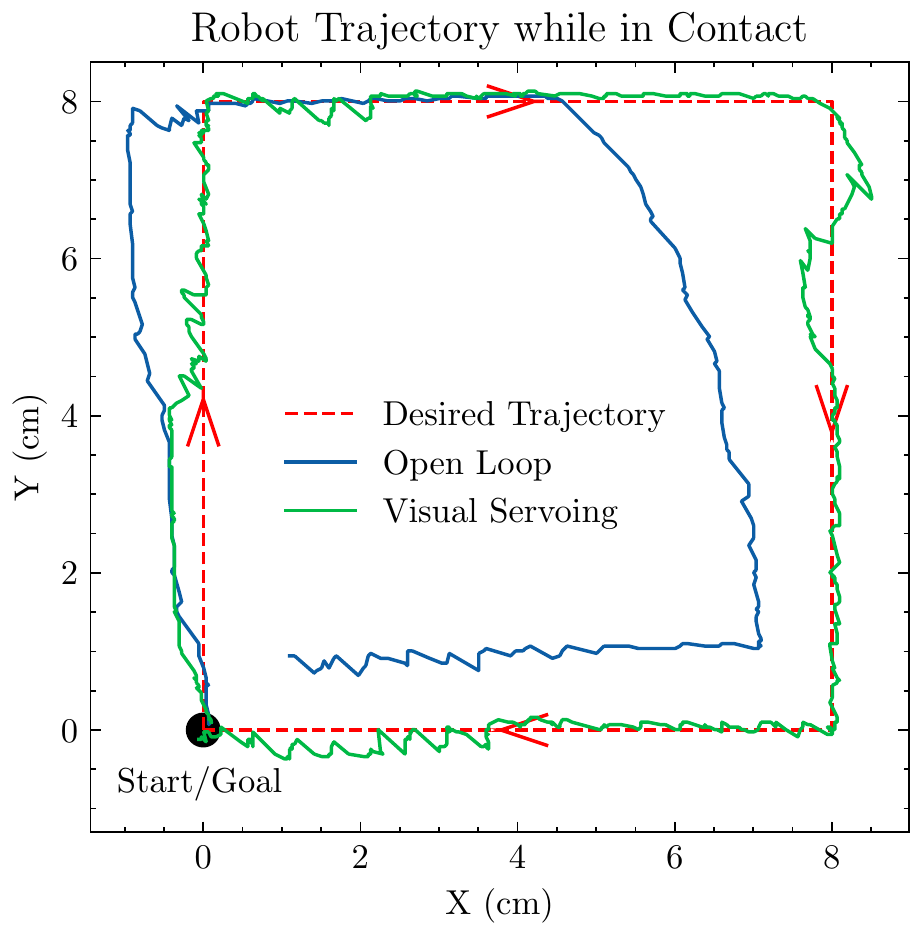}
  \vspace{0mm}
\end{minipage}
\hfill
\begin{minipage}[t]{0.29\textwidth}
% \begin{figure}
  \centering
  \vspace{3mm}
  \includegraphics[width=0.99\linewidth]{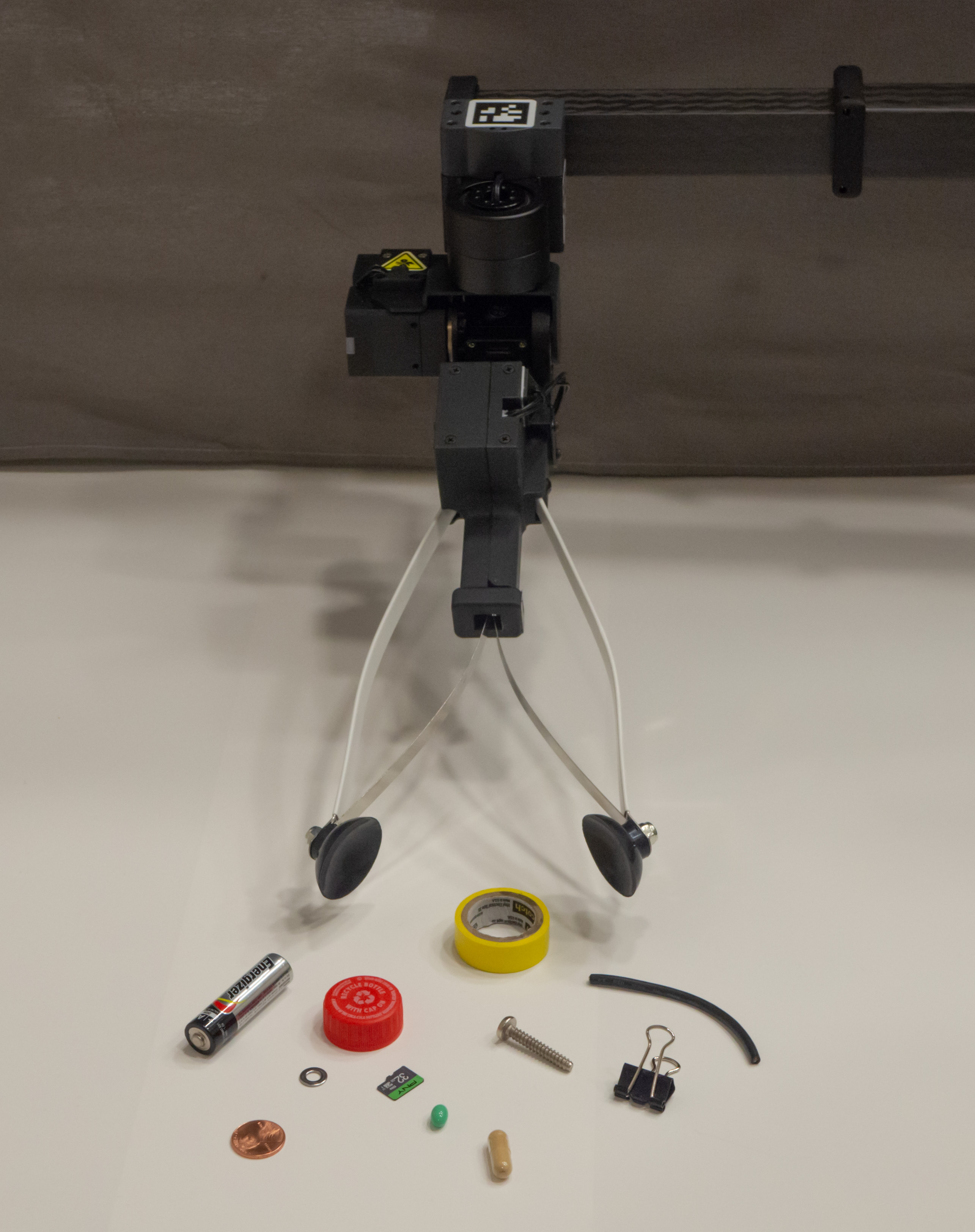}
  \vspace{0mm}
% \end{figure}
\end{minipage}
\caption{\textbf{Left}: The fingertips of the tendon-actuated gripper deflect 4cm when subjected to 5N of lateral force due to deformation of the gripper. \textbf{Middle}: When commanded to trace a square path while in contact with a flat surface (red), a gripper using open-loop control accumulates significant error (blue). Feedback control using image space pressure estimates reduces tracking error (green) \textbf{Right}: In left to right order, this image shows the objects from our grasping evaluation: AA battery, penny, washer, bottle cap, microSD card, small green pill, tape roll, large pill, screw, cable segment, and binder clip.}
\label{fig:robot_multifig}
\end{figure*}
\section{Robotic Control of Pressure} \label{sec:robotic_system}

We evaluated \network{} with robotic manipulation tasks involving pressure objectives and precision grasping. We first show that \network{} can be used to modulate the pressure applied to a surface and increase the spatial accuracy of a compliant robot. We then show how our approach can be used to pick up small objects (penny, screw) that require precision manipulation. For all experiments, we used a Stretch RE1 mobile manipulator with its stock gripper.

%As these items are small and lightweight, the grasping algorithm must be precise. If the gripper is pushed into the surface with too much force, the gripper may bind and fail to close, or the robot actuators may exceed their force limits. If contact is not made with the surface, due to the thin nature of the objects, the robot will fail to pick them up.

% We develop a visual servoing approach with \network{} that succeeds at grasping these small objects. 

\subsection{Making Contact with a Desired Pressure}\label{sec:force_servo}

\network{} can be used to regulate the amount of force a gripper exerts while in contact with a surface. We perform an experiment where the robot is commanded to exert a specified amount of normal force by lowering the gripper to make contact with a pressure sensor.

The pressure estimated by \network{} is integrated with respect to area on the surface to acquire a total force estimate (Eqn. \ref{eqn:pressure_to_force}). The robot uses a simple bang-bang controller to modulate force with the surface by adjusting the height of the gripper using the Stretch RE1's lift joint.

\begin{equation}
\hat{F} = \int f(I)dA
\label{eqn:pressure_to_force}
\end{equation}

% We perform an experiment in which a robot is commanded by a closed-loop bang-bang controller using \network{} to produce an estimate of the normal force on a flat surface (Figure~\ref{fig:force_actual}). This force estimate is compared to the force measured by a pressure sensor. Ground truth force is attained by integrating the pressure along a the surface of the sensor, and estimated force is attained by integrating the pressure along a transformation of the input image.

Each trial begins with the robot placed 3-5 cm above the pressure sensor with \network{} running on a single camera at a rate of 12 Hz.
%The robot is then commanded to move downward onto the pressure sensor until 
Once the pressure estimates indicate that the target force has been achieved, the ground truth force is measured with the pressure sensor. We conduct a total of 60 trials, with 10 trials being recorded for each force level ranging from $0$ to $5$N (Figure \ref{fig:force_actual}).

\network{} can accurately estimate force at higher levels. However, it tends to underestimate forces in the range of $1$ to $2$N, near the boundary of contact. This may be due to differences between the manually operated gripper used for training and the robotic gripper used for testing.

\subsection{Following a Spatial Pressure Trajectory} \label{sec:contact_servo}

Due to the inherent compliance of soft grippers, the precise pose of the gripper can be difficult to control. This is especially true when in contact with a surface, as deformation and friction with the surface can cause the gripper to stick and slip. Figure~\ref{fig:robot_multifig}a shows that the gripper has significant deflection in response to an external disturbance. Additionally, to move the gripper laterally, the Stretch RE1 drives on a carpeted floor with its differential drive mobile base, which can result in movement variations and inaccurate positioning due to wheel slip and other phenomena. 

%Additionally, the actuators of the robot may have backlash or inaccuracy.

% The robot's telescoping arm also has backlash that can interfere .

% \begin{figure}
%   \centering
%   \vspace{2mm}
%   \includegraphics[width=1.0\linewidth]{images/square_inline_arrow.pdf}
%   \vspace{-8mm}
%   \caption{When commanded to trace a square path while in contact (red), a gripper using open-loop control may accumulate significant error (blue) and does not return to the origin. This is due to deformation in the gripper and backlash in the robot kinematics. By adding a simple feedback controller using image space pressure estimates (green), the location of the center of pressure can be accurately controlled. The actual trajectories are measured using the pressure sensor.}
%   \label{fig:tracing_path}
%   \vspace{-2mm}
% \end{figure}

%Additionally, the Stretch RE1 robot uses its mobile base to drive on a carpeted floor for movement in the $X$ direction, and the actuators of the robot have significant backlash.

We show that an open-loop controller accumulates significant error while executing a trajectory in contact (Figure~\ref{fig:robot_multifig}b, blue). The robot gripper was rotated and lowered to a constant height such that one fingertip was in contact with the surface.
%\jeremy{was there a starting force for open loop/target force for closed loop? might be useful to know how much friction there was} \patrick{When held at a fixed height, the force changed a lot depending on if you were pulling or pushing the gripper forwards or backwards}
The robot was then commanded to move in an $8$cm square path. The true path of the gripper was measured by calculating the center of pressure detected by the pressure sensor. 

To achieve a higher accuracy, we use an image-based visual servoing (IBVS) controller \cite{hutchinson1996tutorial} that leverages the image space pressure estimates from \network{} (Figure~\ref{fig:robot_multifig}b, green). The error function $E(t)$ uses the position of the maxima in the estimated pressure image, $M(t)$, and a desired target position in image space $T$.

\begin{equation}
    E(t) = \begin{bmatrix}e_x\\e_y\end{bmatrix}=\begin{bmatrix}T_x-M_x(t)\\T_y-M_y(t)\end{bmatrix}
\end{equation}

This error is transformed into robot actuator commands $\dot{q}$ with the image Jacobian $J$ and a gain $\lambda$: $\dot{q}(t) = \lambda J^{+}E(t)$, where $J^{+}$ is the pseudo-inverse. Because we observe both the target and gripper contact location in the same image, our controller is `endpoint closed-loop'~\cite{hutchinson1996tutorial} and robust to inaccuracies in $J$. %\patrick{Add MAE metrics}

\begin{figure*}
  \centering
  \vspace{2mm}
  \includegraphics[width=1.0\linewidth]{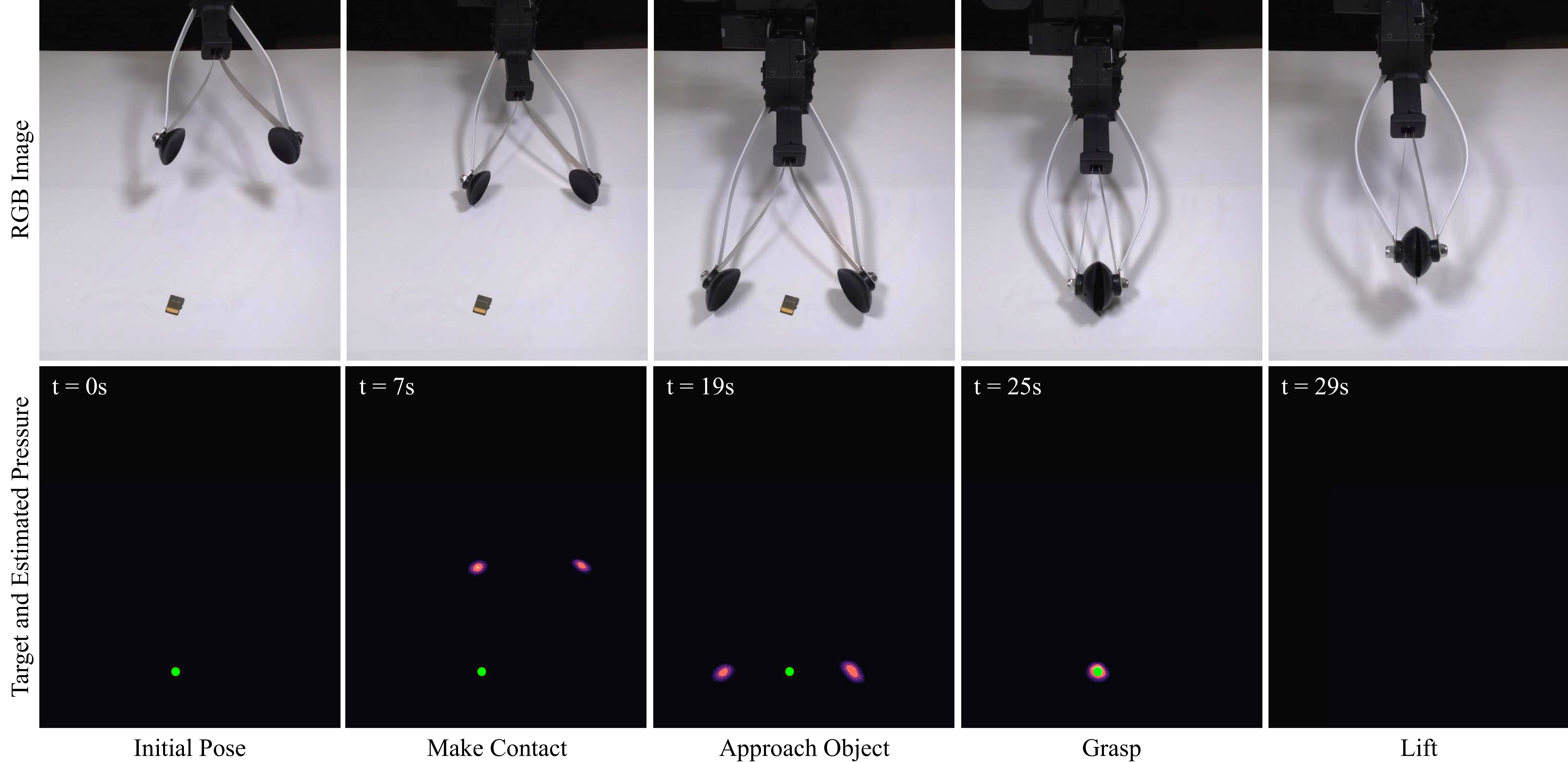}
  \vspace{-5mm}
  \caption{\textbf{Left to Right}: Grasping a 1mm thick microSD card. \textbf{t=0s}: The centroid of the object shown as a green circle is estimated using the RGB input image. \textbf{t=7s}: The robot makes contact with the uninstrumented tabletop to achieve a desired pressure and estimate its location. Fingertip contact results in two ellipsoidal contact pressure regions. \textbf{t=19s}: The robot moves the estimated fingertip pressure regions to the centroid of the object. \textbf{t=25s and t=29s}: The gripper closes while maintaining a desired pressure on the surface to grasp and then pick up the microSD card.}
  
  %The robot is able to pick up small objects using a simple algorithm: it presses into the surface and uses the pressure predictions from \network{} to steer the contact points onto the object.  
  \label{fig:pick_up_object}
  \vspace{-2mm}
\end{figure*}

\subsection{Grasping Small Low-Profile Objects}

To demonstrate the real-world value of \method{}, we perform grasping trials with a range of objects (Figure~\ref{fig:robot_multifig}c), including very thin objects. Humans typically grasp these small objects by using their fingers to first make contact with the surface near the object, then slide their fingertips to close around the object. We take inspiration from this approach and design a robot control algorithm to grasp objects while maintaining contact with the surface.

The robot must autonomously approach the object, grasp it, and pick it up without dropping it for 5 seconds (Figure~\ref{fig:pick_up_object}). Trials where any of the robot's actuators exceed their torque limits are marked as failures. We remove the pressure sensor from our capture rig (Figure \ref{fig:camera_cage}) during grasping experiments, providing the robot with a larger workspace and demonstrating that \network{} can generalize to a new surface. We conduct 10 grasping attempts for each object. The object is reset to a random position and orientation after each trial. 

Our system used a simple color thresholding algorithm to find the centroid of the object in the image. The robot starts with the gripper positioned above the surface and is lowered until pressure above a threshold is reported by \network{}. The normal force exerted by the gripper is continuously controlled to maintain a set force (Sec \ref{sec:force_servo}).

We then perform visual servoing (Sec \ref{sec:contact_servo}) to grasp the object. Our algorithm attempts to navigate the mean position of the two local pressure maxima produced by the gripper fingertips to the object centroid in image space.
%The position of the gripper in image space is estimated by calculating the mean position of the two local maxima produced by the gripper fingertips. A simple control loop (Eqn X) is used to servo the gripper position on top of the object centroid while maintaining contact.
Once the average fingertip position is within a fixed radius of the object centroid, the gripper is closed and lifted.

\begin{table}[b]
\centering
% use resizebox if table gets too big, also uncomment the } after \end{tabular}
%\resizebox{\textwidth}{!}{
\begin{tabular}{c|c|c}
    \textbf{Object} & \textbf{Dims. L$\times$W$\times$H} & \textbf{Grasp Successes/Trials}\\\hline
    Washer & 10$\times$10$\times$1 mm & 9/10 \\\hline
    Small Green Pill & 10$\times$6$\times$6 mm  & 10/10  \\\hline
    Large Pill & 21$\times$8$\times$8 mm  & 9/10 \\\hline
    MicroSD Card & 15$\times$11$\times$1 mm  & 8/10 \\\hline
    Cable Segment & 82$\times$4$\times$4 mm  & 10/10 \\\hline
    Penny & 19$\times$19$\times$1.5 mm  & 9/10 \\\hline
    Bottle Cap & 30$\times$30$\times$13 mm  & 9/10 \\\hline
    AA Battery & 50$\times$14$\times$14 mm  & 9/10 \\\hline
    Binder Clip & 25$\times$24$\times$19 mm  & 9/10 \\\hline
    Screw & 32$\times$9$\times$9 mm  & 10/10 \\\hline
    Tape Roll & 36$\times$36$\times$13 mm  & 10/10 \\\hline
\end{tabular}
%}
\caption{Object Dimensions and Grasping Results
\label{tab:grasping_results}}
\end{table}

\subsection{Grasping Results} We find that \method{} allows the robot to accomplish precision grasping using images from a single RGB camera. The robot is able to grasp all 11 objects in our set (Table~\ref{tab:grasping_results}), and achieves an average success rate of 93\%. The robot is also able to maintain contact with the surface, allowing the visual servoing controller to accurately track the position of the robot in image space and modulate normal force to grasp thin objects on a flat surface.

Failures during grasping experiments can be attributed to a few causes. As our dataset was collected without distractor objects present, when objects are placed in the camera's field of view, the network occasionally estimates pressure near the object in image space. This extra pressure estimate may cause the gripper to lift off the surface. We also find that the network may occasionally overestimate the gripper pressure, also causing the gripper to be in inconsistent contact with the surface. In very rare cases, pressure is underestimated, causing the gripper to be driven into the surface such that the motor torque limits are exceeded and the trial is stopped. We would expect additional training data to increase robustness and alleviate these issues.

\section{Conclusion}

We present \method{}, a method to visually estimate pressure from changes in the appearance of a soft gripper. We demonstrate that a trained model can accurately estimate pressure for two designs: a tendon-actuated gripper and a pneumatic gripper. These pressure estimates can be used to perform closed-loop control of a robot to maintain a desired pressure, accurately trace a trajectory, and successfully manipulate small objects. Our results suggest that visual estimation of pressure is a promising approach for soft robotic grippers.

%We hope that this provides a promising direction for future work in estimating pressure from images.

\bibliographystyle{IEEEtran}
\bibliography{cited}

\begin{thebibliography}{10}
\providecommand{\url}[1]{#1}
\csname url@rmstyle\endcsname
\providecommand{\newblock}{\relax}
\providecommand{\bibinfo}[2]{#2}
\providecommand\BIBentrySTDinterwordspacing{\spaceskip=0pt\relax}
\providecommand\BIBentryALTinterwordstretchfactor{4}
\providecommand\BIBentryALTinterwordspacing{\spaceskip=\fontdimen2\font plus
\BIBentryALTinterwordstretchfactor\fontdimen3\font minus
  \fontdimen4\font\relax}
\providecommand\BIBforeignlanguage[2]{{%
\expandafter\ifx\csname l@#1\endcsname\relax
\typeout{** WARNING: IEEEtran.bst: No hyphenation pattern has been}%
\typeout{** loaded for the language `#1'. Using the pattern for}%
\typeout{** the default language instead.}%
\else
\language=\csname l@#1\endcsname
\fi
#2}}

\bibitem{grady2022pressurevision}
P.~Grady, C.~Tang, S.~Brahmbhatt, C.~D. Twigg, C.~Wan, J.~Hays, and C.~C. Kemp,
  ``{PressureVision:} estimating hand pressure from a single {RGB} image,''
  \emph{European Conference on Computer Vision (ECCV)}, 2022.

\bibitem{kazemi2014human}
M.~Kazemi, J.-S. Valois, J.~A. Bagnell, and N.~Pollard, ``Human-inspired force
  compliant grasping primitives,'' \emph{Autonomous Robots}, vol.~37, no.~2,
  pp. 209--225, 2014.

\bibitem{eppner2015exploitation}
C.~Eppner, R.~Deimel, J.~Alvarez-Ruiz, M.~Maertens, and O.~Brock,
  ``Exploitation of environmental constraints in human and robotic grasping,''
  \emph{The International Journal of Robotics Research}, vol.~34, no.~7, pp.
  1021--1038, 2015.

\bibitem{ciocarlie2014velo}
M.~Ciocarlie, F.~M. Hicks, R.~Holmberg, J.~Hawke, M.~Schlicht, J.~Gee,
  S.~Stanford, and R.~Bahadur, ``The {V}elo gripper: A versatile
  single-actuator design for enveloping, parallel and fingertip grasps,''
  \emph{The International Journal of Robotics Research}, vol.~33, no.~5, pp.
  753--767, 2014.

\bibitem{babin2018picking}
V.~Babin and C.~Gosselin, ``Picking, grasping, or scooping small objects lying
  on flat surfaces: A design approach,'' \emph{The International Journal of
  Robotics Research}, vol.~37, no.~12, 2018.

\bibitem{yoon2021analysis}
D.~Yoon and Y.~Choi, ``Analysis of fingertip force vector for pinch-lifting
  gripper with robust adaptation to environments,'' \emph{IEEE Transactions on
  Robotics}, vol.~37, no.~4, pp. 1127--1143, 2021.

\bibitem{eppner2017visual}
C.~Eppner and O.~Brock, ``Visual detection of opportunities to exploit contact
  in grasping using contextual multi-armed bandits,'' in \emph{2017 IEEE/RSJ
  International Conference on Intelligent Robots and Systems (IROS)}.\hskip 1em
  plus 0.5em minus 0.4em\relax IEEE, 2017, pp. 273--278.

\bibitem{pozzi2020hand}
M.~Pozzi, S.~Marullo, G.~Salvietti, J.~Bimbo, M.~Malvezzi, and D.~Prattichizzo,
  ``Hand closure model for planning top grasps with soft robotic hands,''
  \emph{The International Journal of Robotics Research}, vol.~39, no.~14, pp.
  1706--1723, 2020.

\bibitem{gupta2016learning}
A.~Gupta, C.~Eppner, S.~Levine, and P.~Abbeel, ``Learning dexterous
  manipulation for a soft robotic hand from human demonstrations,'' in
  \emph{2016 IEEE/RSJ International Conference on Intelligent Robots and
  Systems (IROS)}.\hskip 1em plus 0.5em minus 0.4em\relax IEEE, 2016, pp.
  3786--3793.

\bibitem{begej1988planar}
S.~Begej, ``Planar and finger-shaped optical tactile sensors for robotic
  applications,'' \emph{IEEE Journal on Robotics and Automation}, vol.~4,
  no.~5, pp. 472--484, 1988.

\bibitem{li2014localization}
R.~Li, R.~Platt, W.~Yuan, A.~ten Pas, N.~Roscup, M.~A. Srinivasan, and
  E.~Adelson, ``Localization and manipulation of small parts using gelsight
  tactile sensing,'' in \emph{2014 IEEE/RSJ International Conference on
  Intelligent Robots and Systems (IROS)}.\hskip 1em plus 0.5em minus
  0.4em\relax IEEE, 2014, pp. 3988--3993.

\bibitem{yamaguchi2016combining}
A.~Yamaguchi and C.~G. Atkeson, ``Combining finger vision and optical tactile
  sensing: Reducing and handling errors while cutting vegetables,'' in
  \emph{2016 IEEE-RAS 16th International Conference on Humanoid Robots}.\hskip
  1em plus 0.5em minus 0.4em\relax IEEE, 2016, pp. 1045--1051.

\bibitem{kuppuswamy2020soft}
N.~Kuppuswamy, A.~Alspach, A.~Uttamchandani, S.~Creasey, T.~Ikeda, and
  R.~Tedrake, ``Soft-bubble grippers for robust and perceptive manipulation,''
  in \emph{2020 IEEE/RSJ International Conference on Intelligent Robots and
  Systems (IROS)}.\hskip 1em plus 0.5em minus 0.4em\relax IEEE, 2020, pp.
  9917--9924.

\bibitem{lepora2021soft}
N.~F. Lepora, ``Soft biomimetic optical tactile sensing with the {T}ac{T}ip: A
  review,'' \emph{IEEE Sensors Journal}, 2021.

\bibitem{kuang2020vibration}
W.~Kuang, M.~Yip, and J.~Zhang, ``Vibration-based multi-axis force sensing:
  Design, characterization, and modeling,'' \emph{IEEE Robotics and Automation
  Letters}, vol.~5, no.~2, pp. 3082--3089, 2020.

\bibitem{koonjul2011measuring}
G.~S. Koonjul, G.~J. Zeglin, and N.~S. Pollard, ``Measuring contact points from
  displacements with a compliant, articulated robot hand,'' in \emph{2011 IEEE
  International Conference on Robotics and Automation (ICRA)}.\hskip 1em plus
  0.5em minus 0.4em\relax IEEE, 2011, pp. 489--495.

\bibitem{wang2020contact}
S.~Wang, A.~Bhatia, M.~T. Mason, and A.~M. Johnson, ``Contact localization
  using velocity constraints,'' in \emph{2020 IEEE/RSJ International Conference
  on Intelligent Robots and Systems (IROS)}.\hskip 1em plus 0.5em minus
  0.4em\relax IEEE, 2020, pp. 7351--7358.

\bibitem{nazari2021image}
A.~A. Nazari, F.~Janabi-Sharifi, and K.~Zareinia, ``Image-based force
  estimation in medical applications: A review,'' \emph{IEEE Sensors Journal},
  vol.~21, no.~7, pp. 8805--8830, 2021.

\bibitem{kennedy2005vision}
C.~W. Kennedy and J.~P. Desai, ``A vision-based approach for estimating contact
  forces: Applications to robot-assisted surgery,'' \emph{Applied Bionics and
  Biomechanics}, vol.~2, no.~1, pp. 53--60, 2005.

\bibitem{noohi2014using}
E.~Noohi, S.~Parastegari, and M.~{\v{Z}}efran, ``Using monocular images to
  estimate interaction forces during minimally invasive surgery,'' in
  \emph{2014 IEEE/RSJ International Conference on Intelligent Robots and
  Systems}.\hskip 1em plus 0.5em minus 0.4em\relax IEEE, 2014, pp. 4297--4302.

\bibitem{kim2019efficient}
D.~Kim, H.~Cho, H.~Shin, S.-C. Lim, and W.~Hwang, ``An efficient
  three-dimensional convolutional neural network for inferring physical
  interaction force from video,'' \emph{Sensors}, vol.~19, no.~16, p. 3579,
  2019.

\bibitem{marban2019recurrent}
A.~Marban, V.~Srinivasan, W.~Samek, J.~Fern{\'a}ndez, and A.~Casals, ``A
  recurrent convolutional neural network approach for sensorless force
  estimation in robotic surgery,'' \emph{Biomedical Signal Processing and
  Control}, vol.~50, pp. 134--150, 2019.

\bibitem{chua2021toward}
Z.~Chua, A.~M. Jarc, and A.~M. Okamura, ``Toward force estimation in
  robot-assisted surgery using deep learning with vision and robot state,'' in
  \emph{2021 IEEE International Conference on Robotics and Automation
  (ICRA)}.\hskip 1em plus 0.5em minus 0.4em\relax IEEE, 2021, pp.
  12\,335--12\,341.

\bibitem{greminger2003modeling}
M.~A. Greminger and B.~J. Nelson, ``Modeling elastic objects with neural
  networks for vision-based force measurement,'' in \emph{IEEE/RSJ
  International Conference on Intelligent Robots and Systems (IROS)},
  vol.~2.\hskip 1em plus 0.5em minus 0.4em\relax IEEE, 2003, pp. 1278--1283.

\bibitem{li2019connecting}
Y.~Li, J.-Y. Zhu, R.~Tedrake, and A.~Torralba, ``Connecting touch and vision
  via cross-modal prediction,'' in \emph{IEEE Conference on Computer Vision and
  Pattern Recognition (CVPR)}, 2019, pp. 10\,609--10\,618.

\bibitem{zapata2020generation}
B.~S. Zapata-Impata, P.~Gil, Y.~Mezouar, and F.~Torres, ``Generation of tactile
  data from 3d vision and target robotic grasps,'' \emph{IEEE Transactions on
  Haptics}, vol.~14, no.~1, pp. 57--67, 2020.

\bibitem{patel2020deep}
K.~Patel, S.~Iba, and N.~Jamali, ``Deep tactile experience: Estimating tactile
  sensor output from depth sensor data,'' in \emph{2020 IEEE/RSJ International
  Conference on Intelligent Robots and Systems (IROS)}.\hskip 1em plus 0.5em
  minus 0.4em\relax IEEE, 2020, pp. 9846--9853.

\bibitem{hutchinson1996tutorial}
S.~Hutchinson, G.~D. Hager, and P.~I. Corke, ``A tutorial on visual servo
  control,'' \emph{IEEE Transactions on Robotics and Automation}, vol.~12,
  no.~5, pp. 651--670, 1996.

\bibitem{fitzpatrick2004power}
P.~M. Fitzpatrick and E.~R. Torres-Jara, ``The power of the dark side: using
  cast shadows for visually-guided touching,'' in \emph{4th IEEE/RAS
  International Conference on Humanoid Robots, 2004.}, vol.~1.\hskip 1em plus
  0.5em minus 0.4em\relax IEEE, 2004, pp. 437--449.

\bibitem{calli2018robust}
B.~Calli and A.~M. Dollar, ``Robust precision manipulation with simple process
  models using visual servoing techniques with disturbance rejection,''
  \emph{IEEE Transactions on Automation Science and Engineering}, vol.~16,
  no.~1, pp. 406--419, 2018.

\bibitem{morgan2021vision}
A.~S. Morgan, B.~Wen, J.~Liang, A.~Boularias, A.~M. Dollar, and K.~Bekris,
  ``Vision-driven compliant manipulation for reliable, high-precision assembly
  tasks,'' \emph{Robotics: Science and Systems, (RSS)}, 2021.

\bibitem{sikka1994tactile}
P.~Sikka, H.~Zhang, and S.~Sutphen, ``Tactile servo: Control of touch-driven
  robot motion,'' in \emph{Experimental Robotics III}.\hskip 1em plus 0.5em
  minus 0.4em\relax Springer, 1994, pp. 219--233.

\bibitem{chen1995edge}
N.~Chen, H.~Zhang, and R.~Rink, ``Edge tracking using tactile servo,'' in
  \emph{Proceedings 1995 IEEE/RSJ International Conference on Intelligent
  Robots and Systems. Human Robot Interaction and Cooperative Robots},
  vol.~2.\hskip 1em plus 0.5em minus 0.4em\relax IEEE, 1995, pp. 84--89.

\bibitem{li2013control}
Q.~Li, C.~Sch{\"u}rmann, R.~Haschke, and H.~J. Ritter, ``A control framework
  for tactile servoing.'' in \emph{Robotics: Science and Systems, (RSS)}.\hskip
  1em plus 0.5em minus 0.4em\relax Citeseer, 2013.

\bibitem{wen2021tactile}
C.-T. Wen, S.~Arai, J.~Kinugawa, and K.~Kosuge, ``Tactile servoing based
  pressure distribution control of a manipulator using a convolutional neural
  network,'' \emph{IEEE Access}, vol.~9, pp. 117\,132--117\,139, 2021.

\bibitem{lepora2021pose}
N.~F. Lepora and J.~Lloyd, ``Pose-based tactile servoing: Controlled soft touch
  using deep learning,'' \emph{IEEE Robotics \& Automation Magazine}, vol.~28,
  no.~4, pp. 43--55, 2021.

\bibitem{kemp2021design}
C.~C. Kemp, A.~Edsinger, H.~M. Clever, and B.~Matulevich, ``The design of
  {S}tretch: {A} compact, lightweight mobile manipulator for indoor human
  environments,'' in \emph{IEEE International Conference on Robotics and
  Automation (ICRA)}, 2022.

\bibitem{softgripping}
\BIBentryALTinterwordspacing
{SoftGripping by Wegard GmbH}. (2022) {SoftGripping}, the modular design system
  for flexible gripping. [Online]. Available: \url{https://soft-gripping.com/}
\BIBentrySTDinterwordspacing

\bibitem{senselmorph}
Sensel, ``Sensel {M}orph haptic sensing tablet,''
  \url{https://morph.sensel.com/}, Last accessed on 2022-02-22.

\bibitem{garrido2014automatic}
S.~Garrido-Jurado, R.~Mu{\~n}oz-Salinas, F.~J. Madrid-Cuevas, and M.~J.
  Mar{\'\i}n-Jim{\'e}nez, ``Automatic generation and detection of highly
  reliable fiducial markers under occlusion,'' \emph{Pattern Recognition},
  vol.~47, no.~6, pp. 2280--2292, 2014.

\bibitem{resnet}
K.~He, X.~Zhang, S.~Ren, and J.~Sun, ``Deep residual learning for image
  recognition,'' in \emph{2016 {IEEE} Conference on Computer Vision and Pattern
  Recognition, (CVPR)}, 2016, pp. 770--778.

\bibitem{squeeze-excitation}
J.~Hu, L.~Shen, and G.~Sun, ``Squeeze-and-excitation networks,'' in \emph{2018
  {IEEE} Conference on Computer Vision and Pattern Recognition, (CVPR)}, 2018.

\bibitem{resnext}
S.~Xie, R.~B. Girshick, P.~Doll{\'{a}}r, Z.~Tu, and K.~He, ``Aggregated
  residual transformations for deep neural networks,'' in \emph{2017 {IEEE}
  Conference on Computer Vision and Pattern Recognition, (CVPR)}, 2017.

\bibitem{segmentation_models_pytorch}
P.~Yakubovskiy, ``Segmentation models pytorch,''
  \url{https://github.com/qubvel/segmentation_models.pytorch}, 2020.

\bibitem{deng2009imagenet}
J.~Deng, W.~Dong, R.~Socher, L.-J. Li, K.~Li, and L.~Fei-Fei, ``Imagenet: A
  large-scale hierarchical image database,'' in \emph{2009 IEEE Conference on
  Computer Vision and Pattern Recognition, (CVPR)}.\hskip 1em plus 0.5em minus
  0.4em\relax IEEE, 2009, pp. 248--255.

\bibitem{fpn}
T.-Y. Lin, P.~Doll{\'a}r, R.~Girshick, K.~He, B.~Hariharan, and S.~Belongie,
  ``Feature pyramid networks for object detection,'' in \emph{IEEE Conference
  on Computer Vision and Pattern Recognition, (CVPR)}, 2017, pp. 2117--2125.

\bibitem{adam}
D.~P. Kingma and J.~Ba, ``Adam: {A} method for stochastic optimization,'' in
  \emph{3rd International Conference on Learning Representations, (ICLR) 2015},
  Y.~Bengio and Y.~LeCun, Eds., 2015.

\end{thebibliography}

\end{document}